\documentclass[runningheads]{llncs}
\usepackage[T1]{fontenc}
\usepackage{graphicx}
\usepackage{hyperref}
\usepackage{float}
\usepackage{booktabs}
\usepackage{amsmath}
\usepackage{multirow}
\usepackage{xcolor}
\usepackage{orcidlink}

\usepackage{amssymb}
\begin{document}

\title{D-TAIA: Domain-Aware LLM Adaptation for Multi-Task Predictive Process Monitoring}
\titlerunning{D-TAIA: Domain-Aware LLM Adaptation for MT-PPM}

\author{Sjoerd van Straten \orcidlink{0009-0002-5772-9073}\inst{1} \and Christine Jacob\inst{1} \and
Marwan Hassani \orcidlink{0000-0002-4027-4351}\inst{1,2}}
\authorrunning{S. van Straten et al.}
\institute{
  Department of Mathematics and Computer Science, Eindhoven University of Technology, The Netherlands\\
  \email{\{h.a.j.v.straten, c.christine.jacob, m.hassani\}@tue.nl}
  \and
  Department of Mathematics and Computer Science, University of Wuppertal, Germany\\
  \email{hassani@uni-wuppertal.de}}

\maketitle{}             
\begin{abstract}

Predictive Process Monitoring (PPM) enables organizations to forecast future process behavior, such as the next activity and remaining time of ongoing cases. In practice, three conditions cause existing methods to degrade, namely data scarcity, high process entropy and distributional shift. While Foundation Models (FMs), especially Large Language Models (LLMs), offer a new paradigm through broad sequential reasoning, adapting them to multi-task PPM under these conditions remains an open challenge. Existing FM-based approaches either lack mechanisms for handling distributional shift or rely on direct regression heads that can be structurally misaligned with continuous time prediction tasks. This paper introduces D-TAIA (Domain-aware Training and Attention-based Inference Architecture), a framework for a joint next activity and remaining time prediction task via parameter-efficient fine-tuning of an FM backbone. Our approach combines domain-aware triplet loss (DATL) pre-training with FAISS-based nearest neighbor retrieval for remaining time prediction, and adopts the TAIA inference strategy to preserve pre-trained sequential reasoning during fine-tuning. Evaluated across four real-world event logs, D-TAIA consistently shows SOTA or competitive performance compared to a fine-tuned LLM and a recurrent neural network baseline. Ablation studies confirm that techniques from NLP and computer vision can be transferred effectively to PPM with only a 10M-parameter backbone, though component contributions vary by dataset entropy.

\end{abstract}

\vspace{-2em}
\section{Introduction} \label{sec:introduction}
Process Mining leverages event logs from business information systems to discover, monitor and optimize workflows~\cite{WVDA_2004}. Predictive Process Monitoring (PPM) predicts future attributes of ongoing process instances, primarily in the form of final outcome, next activity (NA) and remaining time (RT) prediction tasks~\cite{ppm_2024}. Accurate multi-task prediction of both NA and RT enables organizations to allocate resources efficiently and intervene proactively in at-risk cases. 

In practice, three conditions cause existing PPM methods to degrade. First, \textit{data scarcity} arises when an event log lacks sufficient cases for a model to learn a reliable process representation. This problem is compounded by class imbalance across rare activities and cross-validation partitioning effects~\cite{peeperkorn_2024}. Second, \textit{high process entropy} quantifies how difficult it is to predict the next activity given the observed prefix due to high trace variability. Formally, the conditional entropy $H(a_{k+1} \mid \pi_k)$ measures the spread of plausible continuations to next activities at location $k+1$ from a given prefix $\pi_k$ at location $k$, where a high value makes prediction inherently harder~\cite{back2019entropy}. Third, \textit{distributional shift} arises when unseen process variants appear at inference time, violating the i.i.d. assumption that underlies standard supervised learning~\cite{darwin_2023}. Each condition alone is addressed by at least one existing method, yet no approach handles all three simultaneously for a joint NA and RT prediction.

Foundation Models (FMs), especially Large Language Models (LLMs), offer a more flexible paradigm than classical deep learning models through broad sequential reasoning pre-trained on text corpora. Oyamada et al.~\cite{oyamada_2025} demonstrate that parameter-efficient fine-tuning (PEFT) with task-specific tokenization can adapt LLMs directly to process data for joint NA and RT prediction, bypassing the need for narrative-style formulation. However, their approach relies on a direct regression head for RT prediction, which inherits the weakness of autoregressive models trained as next-token classifiers for continuous numeric output. Furthermore, their framework includes no mechanism for handling data scarcity, high process entropy or unseen trace variants at inference time, which are challenges that PPM methods most frequently fail to address in practice. 

This paper introduces D-TAIA (Domain-aware Training and Attention-based Inference Architecture), a framework for joint NA and RT prediction using parameter-efficient fine-tuning of FMs. D-TAIA combines an LLM backbone with LoRA~\cite{lora_2021} fine-tuning to minimize computational overhead. Domain-aware triplet loss (DATL)~\cite{datl_2023} pre-training with FAISS-based nearest neighbor retrieval~\cite{faiss_2025} is used for improved numerical RT prediction, targeting data scarcity and high process entropy. The TAIA~\cite{TAIA_2024} inference strategy preserves pre-trained sequential reasoning during fine-tuning on small logs, addressing distributional shift architecturally.

The contributions of this paper are threefold. (i) we show that non-parametric time estimation via FAISS-based nearest neighbor retrieval addresses the regression weakness of autoregressive LLMs more than direct prediction, particularly on high-entropy logs; (ii) we combine DATL pre-training with attention-based inference for joint PPM under data scarcity and high process entropy; and (iii) we evaluate our proposed framework across four real-world event logs of varying complexity, showing that it matches or improves upon a fine-tuned LLM and recurrent neural network (RNN) baseline, with the clearest margins at short prefix lengths and on the highest entropy log.

The remainder of this paper is structured as follows. Section~\ref{sec:preliminaries_problem_definition} covers preliminaries and formally introduces the problem definition. Section~\ref{sec:related_work} explains the related work and defines the research gap. Section~\ref{sec:methodology} describes our framework, while Section~\ref{sec:experimental_setup} outlines the experimental setup. Section~\ref{sec:evaluation_results} extensively analyzes the results, and Section~\ref{sec:conclusion-future_work} concludes with a discussion, limitations and avenues for future work. 
\vspace{-1em}
\section{Preliminaries and Problem Definition} \label{sec:preliminaries_problem_definition}

Let $\mathcal{A}$ denote the universe of all process activities. The execution of a process instance yields a trace, defined as a finite sequence $\sigma = \langle e_1, e_2, \dots, e_L \rangle$ of length $L$, where each event $e_i = (a_i, t_i)$ consists of an activity label $a_i \in \mathcal{A}$ and a timestamp $t_i$. An \textit{event log} $\mathcal{L} = \{\sigma_1, \sigma_2, \ldots, \sigma_N\}$ is a finite collection of such traces. Given a trace $\sigma$, the \textit{prefix} of length $k$ is defined as $\pi_k(\sigma) = \langle e_1, e_2, \ldots, e_k \rangle$. At prediction time only $\pi_k(\sigma)$ is observable, the \textit{suffix} $\langle e_{k+1}, \ldots, e_L \rangle$ remains unknown. In this work, we address two prediction tasks over such prefixes, namely \textit{next activity prediction} and \textit{remaining time prediction}.

\begin{definition}[Next Activity Prediction]\label{def:nap} Given a prefix $\pi_k(\sigma)$ of length $1 \leq k < L$, next activity prediction is the task of predicting activity $\hat{a}_{k+1}$ that should ideally match the ground truth target $y = a_{k+1}$.
\end{definition}

\begin{definition}[Remaining Time Prediction] Given a prefix $\pi_k(\sigma)$ with final observed event at time $t_k$ and a completed trace $\sigma$ with final event at time $t_L$, remaining time prediction is the task of estimating $\mathrm{RT}(\sigma, k) \;=\; t_L - t_k$.\end{definition}

The central objective of this work is the joint prediction of both NA and RT tasks from a single model $f_\theta$. Three conditions compound the difficulty of this objective:

\begin{itemize}
    \item \textbf{Data Scarcity.} Data scarcity arises when $\lvert \mathcal{L} \rvert$ is insufficient for $f_\theta$ to learn a reliable process representation, a problem compounded by class imbalance across rare activities $a \in \mathcal{A}$ and cross-validation partitioning effects~\cite{peeperkorn_2024}.
    \item \textbf{High Process Entropy.} Process entropy, formally defined as $H(a_{k+1} \mid \pi_k) = -\sum_{a \in \mathcal{A}} p(a \mid \pi_k) \log p(a \mid \pi_k)$, measures the spread of plausible continuations from a given prefix. A high value makes NA prediction especially difficult~\cite{back2019entropy}. Section~\ref{sec:methodology} uses a computationally efficient proxy for this quantity, based on the entropy of the activity distribution observed \textit{within} each prefix, for domain binning.
    \item \textbf{Distributional Shift.} This arises when the test distribution $P_{\text{test}}(a_{k+1} \mid \pi_k) \neq P_{\text{train}}(a_{k+1} \mid \pi_k)$, violating the i.i.d. assumption underlying standard supervised learning~\cite{darwin_2023}. 
\end{itemize} 
\vspace{-1em}
\section{Related Work}
\label{sec:related_work}

\subsubsection{Predictive Process Monitoring (PPM) Approaches.} Non-FM methods each target a single condition and, in most cases, a single task (cf. Table~\ref{tab:related_work}). Weinzierl et al.~\cite{weinzierl_2025} address data scarcity via cross-organizational transfer learning, but require a structurally related source log and predict outcome only. High process entropy is targeted separately by two methods. Zare et al.~\cite{zare_2025} incorporate entropy directly into attention weighting for NA prediction, while the graph-based PGTNet~\cite{amiri_elyasi_2024} achieves strong RT prediction under high entropy by encoding control-flow as graph structure, at the cost of degraded performance on sparse logs. Online continual learning approaches~\cite{van_straten_2026,hassani2026chameleons} handle concept drift for NA prediction in a streaming setup, but omit RT prediction entirely.

Within FM-based PPM, two directions dominate. Prompt-based approaches~\cite{padella_2025} exploit pre-trained semantic knowledge to generalize under data scarcity without gradient updates, but target total case duration rather than the joint NA and RT tasks considered here. Rebmann et al.~\cite{rebmann2024evaluating} attribute this limitation to in-context learning itself being insufficient for NA prediction. Oyamada et al.~\cite{oyamada_2025} instead show that parameter-efficient fine-tuning with task-specific tokenization outperforms both RNN and prompt-based baselines for joint NA and RT prediction, but estimates RT via a single point estimate head rather than a retrieval-weighted approach, and provide no mechanism for high entropy or distributional shift. Berti \& van der Aalst~\cite{icfm_2025} propose a mixture-of-experts (MoE) FM with in-context adaptation, providing partial OOD coverage at the cost of vast heterogeneous pre-training data. Narrative text-serialization has also been applied to related PPM tasks such as suffix prediction~\cite{lupin_2024}, which reports accuracy gains concentrated on high-variant logs but does not target data scarcity or distributional shift, and outcome prediction~\cite{pasquadibisceglie2025leveraging}, which targets neither condition.

\begin{table}[t]
\centering
\caption{Overview of related work and the research gap. Axes: \textbf{LD} = Low volume of data; \textbf{HE} =
High-Entropy robustness; \textbf{OOD} = Out-of-Distribution generalization
within a log; \textbf{MT} = Multi-task NA and RT prediction.
\checkmark~= explicitly addressed with a dedicated mechanism and
supporting experiment; $\sim$~= a mechanism is present or claimed but not
directly evaluated.}
\label{tab:related_work}
\resizebox{\textwidth}{!}{%
\begin{tabular}{llccccc}
\toprule
\textbf{Authors} & \textbf{Mechanism}
  & \textbf{Task} & \textbf{LD} & \textbf{HE} & \textbf{OOD}
  & \textbf{MT} \\
\midrule
\multicolumn{7}{l}{\textit{PPM -- non-FM}} \\
\midrule
Weinzierl et al. (2025)~\cite{weinzierl_2025}       & Cross-organization transfer learning       & Outcome      & $\sim$ &        &        &  \\
Amiri Elyasi et al. (2024)~\cite{amiri_elyasi_2024}  & Graph-based control-flow encoding          & RT      &        & \checkmark &   &  \\
Zare et al. (2025)~\cite{zare_2025}                  & Entropy-weighted attention                 & NA      &        & \checkmark &   &  \\
van Straten (2026); Hassani et al. (2026)~\cite{van_straten_2026,hassani2026chameleons} & Continual learning under concept drift & NA & $\sim$ & & & \\
Rama-Maneiro et al. (2023)~\cite{rama2023deep}   & Benchmark of 10 non-FM approaches          & NA, Suffix, RT &        &        &        & $\sim$ \\
\midrule
\multicolumn{7}{l}{\textit{PPM -- FM-based}} \\
\midrule
Padella et al. (2025)~\cite{padella_2025}            & Prompting, no gradient update             & Duration & \checkmark &        &        &  \\
Rebmann et al. (2024)~\cite{rebmann2024evaluating}   & Fine-tuning (ICL found insufficient)      & NA      & $\sim$     &        &        &  \\
Oyamada et al. (2025)~\cite{oyamada_2025}            & PEFT with task-specific tokenization      & NA + RT & $\sim$     &        &        & \checkmark \\
Pasquadibisceglie et al. (2024)~\cite{lupin_2024}    & Multi-view narrative + one-step suffix    & Suffix  &            & $\sim$ &        &  \\
Berti \& van der Aalst (2026)~\cite{icfm_2025}       & MoE + in-context adaptation               & NA + RT & \checkmark &        & $\sim$ & \checkmark \\
\midrule
\multicolumn{7}{l}{\textit{NLP/CV}} \\
\midrule
Jiang et al.\ (2024)~\cite{TAIA_2024}    & Attention/FFN asymmetry at inference       & Classification      & \checkmark &        & \checkmark &  \\
Hu et al.\ (2021)~\cite{lora_2021}       & Low-rank fine-tuning adapters              & Fine-tuning              & $\sim$     &        &            &  \\
Guo \& Lovell (2024)~\cite{datl_2023}    & Domain-aware triplet loss             & Classification     &            & \checkmark & \checkmark &  \\
\midrule
\textbf{D-TAIA (this work)} & \textbf{DATL + FAISS retrieval + TAIA inference} & \textbf{NA + RT} & \checkmark & \checkmark & $\sim$ & \checkmark \\
\bottomrule
\end{tabular}%
}
\vspace{-1.5em}
\end{table}

\vspace{-0.5em}
\subsubsection{Computer Vision and NLP Approaches.} Addressing the computational cost of fine-tuning large FMs, Low-Rank Adaptation (LoRA)~\cite{lora_2021} constrains weight updates to a low-rank bypass by replacing a full update matrix with the product $BA$, where $B \in \mathbb{R}^{d \times r}$ and $A \in \mathbb{R}^{r \times m}$, where $d$ and $m$ are the input and output dimensions of the original weight matrix and $r \ll \min(d, m)$ is the rank of the decomposition. LoRA reduces trainable parameters by orders of magnitude with no additional inference latency, which is convenient for FM adaptation. Building on efficient fine-tuning, the TAIA inference strategy~\cite{TAIA_2024} identifies a functional asymmetry between Transformer components. Feed-Forward Network (FFN) layers absorb distribution-specific knowledge during fine-tuning and overwrite pre-trained memory, whereas attention layers encode transferable sequential reasoning that remains stable across domains. Discarding FFN updates at inference therefore recovers OOD robustness without additional hyperparameters. For domain-invariant representation learning, IRM~\cite{IRM2020} provides a theoretical grounding for learning representations stable across training environments, which DATL~\cite{datl_2023} operationalizes by constructing an embedding space that clusters by semantic behavior rather than domain artifact, making it suitable for retrieval over unseen process variants. Furthermore, FAISS~\cite{faiss_2025} provides an efficient nearest-neighbor index over such embedding spaces, enabling retrieval of the most similar training samples at inference time.

\vspace{-0.5em}
\subsubsection{Research Gap.} As summarized in Table~\ref{tab:related_work}, no existing method simultaneously addresses data scarcity, high process entropy and OOD generalization for joint NA and RT prediction. PPM-specific approaches either target a subset of these challenges or omit one of the two prediction tasks. FM-based methods that do support joint prediction lack mechanisms for handling distributional shift and estimate RT via a single point-estimate head rather than a retrieval-weighted approach. This gap motivates a framework that targets all three conditions. We combine domain-aware representation learning, retrieval-based time estimation and an inference strategy designed for robustness under distributional shift.  
\vspace{-0.5em}
\section{Method}
\label{sec:methodology}

\begin{figure}
    \centering
    \includegraphics[width=1\linewidth]{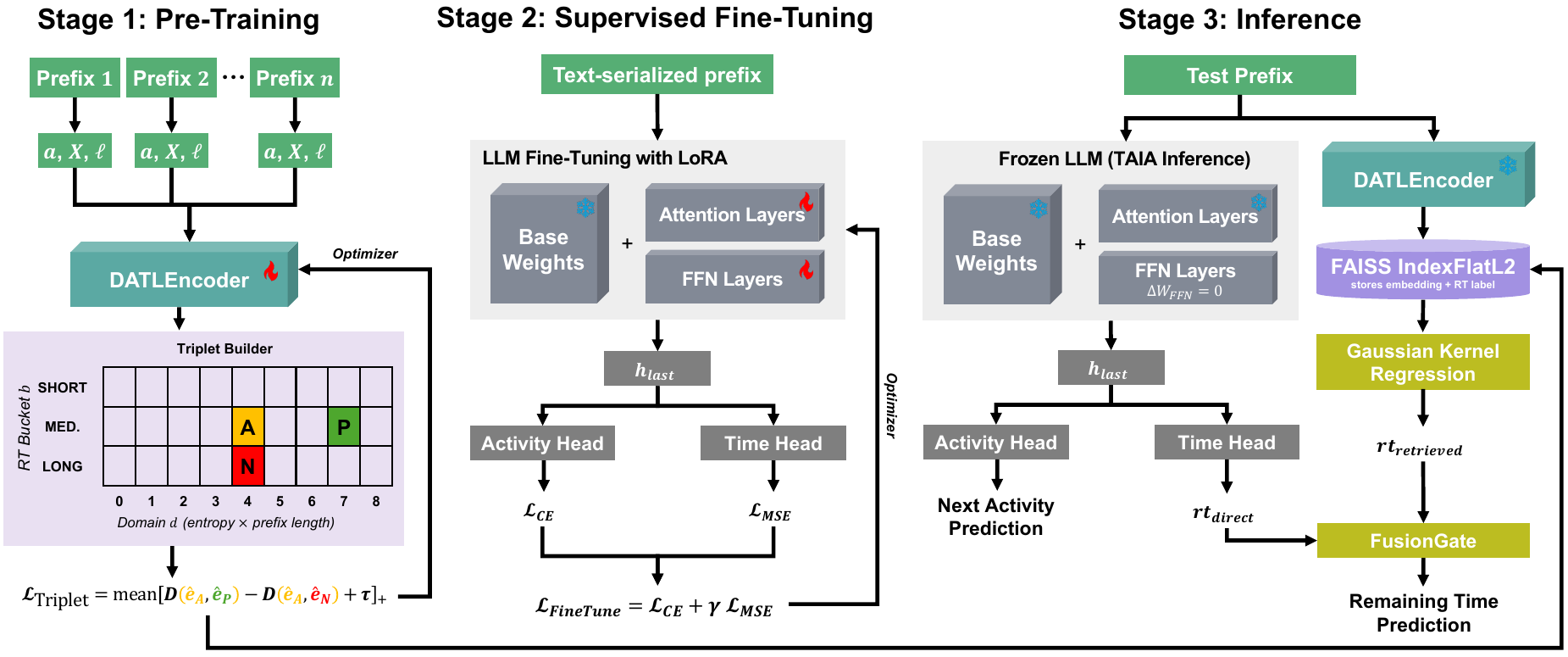}
    \caption{Overview of the D-TAIA framework.}
    \label{fig:architecture}
\end{figure}

We propose D-TAIA (Domain-aware Training and Attention-based Inference Architecture), a three-stage framework for jointly predicting NA and RT
given a prefix (cf. Fig.~\ref{fig:architecture}). NA is predicted by a fine-tuned LLM with a dedicated activity head. RT is predicted by fusing the LLM's time head output with a retrieval estimate obtained by querying a FAISS index of domain-aware prefix embeddings from the training data.\\

\vspace{-0.5em}
\noindent \textbf{Data Preprocessing.} The event logs are filtered to cases with at least two events and sorted chronologically. From each case, every prefix $\pi_k(\sigma)$ with $k \in \{2, \dots, L-1\}$ is paired with the next activity $a_{k+1}$ and the remaining time $RT_k = t_{L} - t_k$. Prefixes are then padded or truncated to $L_{max}=20$ events. Each event carries $F=19$ numerical features. These cover calendar fields, cyclical hour encodings, activity-level duration statistics, workload and case dynamics such as velocity and acceleration. Continuous features and $RT_k$ are Z-score normalized, with statistics frozen on the training set and the normalization inverted at inference.\\

\vspace{-0.5em}
\noindent \textbf{Prefix Labeling and Input Representations.} Each prefix receives two discrete labels used for triplet mining. The domain label $d \in \{0, \dots, 8\}$ crosses three activity-entropy bins with three prefix-length bins, using quantile binning. This entropy is computed over the distribution of activity labels observed within the prefix itself. The RT bucket $b \in \{0, 1, 2\}$ marks the remaining time as short, medium or long via quantile binning at $q_{33}$ and $q_{66}$. All thresholds are computed once on the training set and frozen. The two models consume the same prefix in different forms. DATLEncoder, adapted from~\cite{datl_2023}, receives a tensor triple $(\mathbf{a}, \mathbf{X} \in \mathbb{R}^{L \times F},\ell)$ of activity indices $\mathbf{a}$, the feature matrix $\mathbf{X}$ and the valid length $\ell$. Our LLM backbone instead receives a structured text serialization. It contains a domain token followed by per-step activity-feature pairs, tokenized to 1024 tokens. This format reuses the backbone's tokenizer unchanged and avoids discretizing the numerical features.\\

\vspace{-0.5em}
\noindent\textbf{Stage 1: Pre-Training.} DATLEncoder maps a prefix into a single fixed-size embedding $\hat{e}$ using a compact Transformer encoder. This embedding summarizes both the control-flow and temporal context of the running case. We adapt the domain-aware triplet loss (DATL) from~\cite{datl_2023} to this setting. A standard triplet loss would cluster prefixes by class regardless of domain, encoding domain-specific artifacts that hurt retrieval on unseen process variants. DATL reverses this. The positive sample shares the anchor's RT bucket but comes from a different domain, while the negative sample shares its domain but a different RT bucket. The embedding space therefore clusters by RT behavior rather than domain (Eq.~\ref{eq:datl}):

\begin{equation}\label{eq:datl}
\mathcal{L}_{Triplet} = \frac{1}{|\mathcal{T}|}
\sum_{(A,P,N) \in \mathcal{T}}
[D(\hat{e}_A, \hat{e}_P) - D(\hat{e}_A, \hat{e}_N) + \tau]_+
\end{equation}

where $\mathcal{T}$ is the set of mined triplets with anchor, positive and negative sample indices ($A$,$P$,$N$), $\hat{e}$ the
DATLEncoder embeddings, $D$ the cosine distance, and $\tau$ the margin. After convergence, all training embeddings are L2-normalized,
$\hat{e} \leftarrow \hat{e} / \|\hat{e}\|_2$ and stored together with their $RT_i$ in a FAISS \texttt{IndexFlatL2}~\cite{faiss_2025}, frozen for Stages 2--3.
This normalization step makes squared L2 distance rank-equivalent to cosine distance up to a monotonic transform, $\hat{e}_A - \hat{e}_B\|_2^2 = 2 - 2\cos(\hat{e}_A, \hat{e}_B)$ for unit-norm vectors, so retrieval at Stage 3 is consistent with the metric used during triplet training.\\

\vspace{-0.5em}
\noindent\textbf{Stage 2: Supervised Fine-Tuning.} The LLM is fine-tuned via low-rank adaptation (LoRA)~\cite{lora_2021}, applied to both the attention and FFN sub-layers. This updates only a small fraction of the backbone's parameters while the base weights stay frozen. The last non-padding hidden state $h_{last}$ feeds two heads. The activity head is a linear classifier trained via Cross-Entropy Loss ($\mathcal{L}_{CE}$). The time head is a gated SiLU FFN, followed by a normalized projection to a scalar, trained with Mean Squared Error loss ($\mathcal{L}_{MSE}$). This split anticipates the inference-time adaptation of Stage 3, following TAIA's finding that attention encodes sequential patterns while the FFN encodes task knowledge~\cite{TAIA_2024}. Both heads are optimized jointly under a weighted sum $\mathcal{L}_{FineTune} = \mathcal{L}_{CE} + \gamma \mathcal{L}_{MSE}$.\\

\vspace{-0.5em}
\noindent\textbf{Stage 3: Inference.} We adapt the TAIA inference strategy of~\cite{TAIA_2024} to PPM. TAIA targets the gap between the fine-tuning distribution and the test-time distribution, not the gap between pre-training and fine-tuning. Under this gap, FFN LoRA updates can overwrite pre-trained FFN memory with knowledge specific to the fine-tuning sample, hurting generalization to shifted test prefixes. Attention updates are less prone to this, since they remain bounded by softmax normalization. All FFN LoRA adapters are therefore set to zero before evaluation ($\Delta W_{FFN} = 0$). This reverts each FFN to its pre-trained weights while keeping the attention updates intact. For RT, the same prefix is embedded by DATLEncoder and used to query FAISS for its $n$ nearest neighbors. The neighbor distances $d_i$ are converted to Gaussian kernel weights $w_i = \exp(-d_i^2)$, giving the retrieval estimate $rt_{retrieved} = \sum_i w_i RT_i / \sum_i w_i$. The two RT estimates ($rt_{direct}$ and $rt_{retrieved}$) are complementary. The direct head generalizes parametrically, while the retrieval estimate draws on historically similar cases. A fixed-weight combination, which we term the FusionGate, merges them via $rt_{final} =
\beta \, rt_{direct} + (1-\beta) \, rt_{retrieved}$. It is inverse-transformed to the original time scale and the weight $\beta$ is fixed rather than learned. 
\vspace{-1em}
\section{Experimental Setup} \label{sec:experimental_setup}

\noindent\textbf{Datasets and Metrics.} This paper uses four publicly available event logs from the BPI Challenges~\footnote{\url{https://data.4tu.nl}} (cf. Table~\ref{tab:datasets}), spanning a range of data availability and process complexity. All experiments report three metrics, with 95\% confidence intervals computed across five seeds. Macro-F1 is the unweighted average of the per-class F1 scores over the activity universe $\mathcal{A}$, $F1_{macro} = \frac{1}{|\mathcal{A}|}\sum_{a \in \mathcal{A}} F1_a$, where $F1_a$ is the standard F1 score for class $a$ computed from predictions $\hat{a}_{k+1}$ against ground truth $a_{k+1}$ (Definition~1). This penalizes models that ignore rare activities. RT error is reported as MAE in days, $MAE = \frac{1}{N}\sum_{i=1}^{N} |\widehat{RT}_i - RT_i|$. Wall-clock running time, reported in hours, captures the computational cost of scaling to larger backbones.\\

\begin{table}[h]
\centering
\caption{Characteristics of the event logs used for evaluation.}
\vspace{-0.5em}
\label{tab:datasets}
\footnotesize
\setlength{\tabcolsep}{2.pt}
\renewcommand{\arraystretch}{0.8}
\begin{tabular}{lcccccc}
\toprule
Dataset & \# Cases & \# Events & \# Activities & Avg. Length & Avg. Duration & Entropy \\
\midrule
BPI2012     & 13{,}087 & 262{,}200     & 24  & 20.04 & 8.62   & 3.688 \\
BPI2017     & 31{,}509 & 1{,}202{,}267 & 26  & 38.16 & 21.90  & 3.786 \\
BPI2015\_2  & 832      & 44{,}354      & 410 & 53.31 & 160.49 & 7.105 \\
BPI2020\_DD & 10{,}500 & 56{,}437      & 17  & 5.37  & 11.67  & 2.827 \\
\bottomrule
\end{tabular}
\end{table}

\noindent\textbf{Baselines.} D-TAIA is benchmarked against two baselines. The first, which we refer to as \textit{FT-LLM}, is an LLM fine-tuned directly with LoRA~\cite{lora_2021} following Oyamada et al.~\cite{oyamada_2025}, including their task-specific tokenization rather than D-TAIA's text serialization. It omits DATL pre-training, FAISS retrieval and TAIA inference. Whenever the backbone is varied in our experiments, FT-LLM is evaluated with the same backbone as D-TAIA in that setting. This isolates the effect of D-TAIA's added components from the effect of backbone capacity. Any gap between the two is attributable to the architecture, not to a larger or more capable underlying model. The second, \textit{MT-RNN}, is a multi-task recurrent neural network
trained from scratch for joint NA and RT prediction, following the multi-task LSTM architecture benchmarked by Rama-Maneiro et al.~\cite{rama2023deep}. It represents the classical deep sequence modeling paradigm in which a shared recurrent encoder feeds task-specific classification and regression heads.\\

\vspace{-0.5em}
\noindent\textbf{Implementation Details.} Code is available on GitHub~\footnote{\url{https://github.com/SvStraten/D-TAIA}}. The train-validation-test split is temporal by case start time and fixed across all seeds ($65/15/20$), which affect only weight initialization and mini-batch order. No case appears in more than one split to prevent data leakage. Each experiment is repeated five times over different seeds with mean and 95\% confidence interval over runs reported. Architectural hyperparameters are fixed across all datasets. LoRA $r{=}16$, $\alpha{=}32$; DATLEncoder $d_{model}{=}256$, 8 heads, 4 layers; domain bins $3{\times}3$ (entropy $\times$ length, quantile); RT buckets at $q_{33}$, $q_{66}$; retrieval $n{=}10$; FusionGate $\beta{=}0.5$. The fine-tuning learning rate and loss weight $\gamma$ are instead selected per dataset via grid search on the validation set, over $\{1\text{e-}4, 5\text{e-}4, 1\text{e-}3\}$ and $\{1.0, 2.0, 5.0\}$ respectively. Experiments have been run on an HPC cluster with NVIDIA Tesla V100 (16GB) GPUs.  
\vspace{-0.5em}
\section{Evaluation Results} \label{sec:evaluation_results}

\subsubsection{Backbone Sensitivity.}

\begin{figure}
    \centering
    \includegraphics[width=1\linewidth]{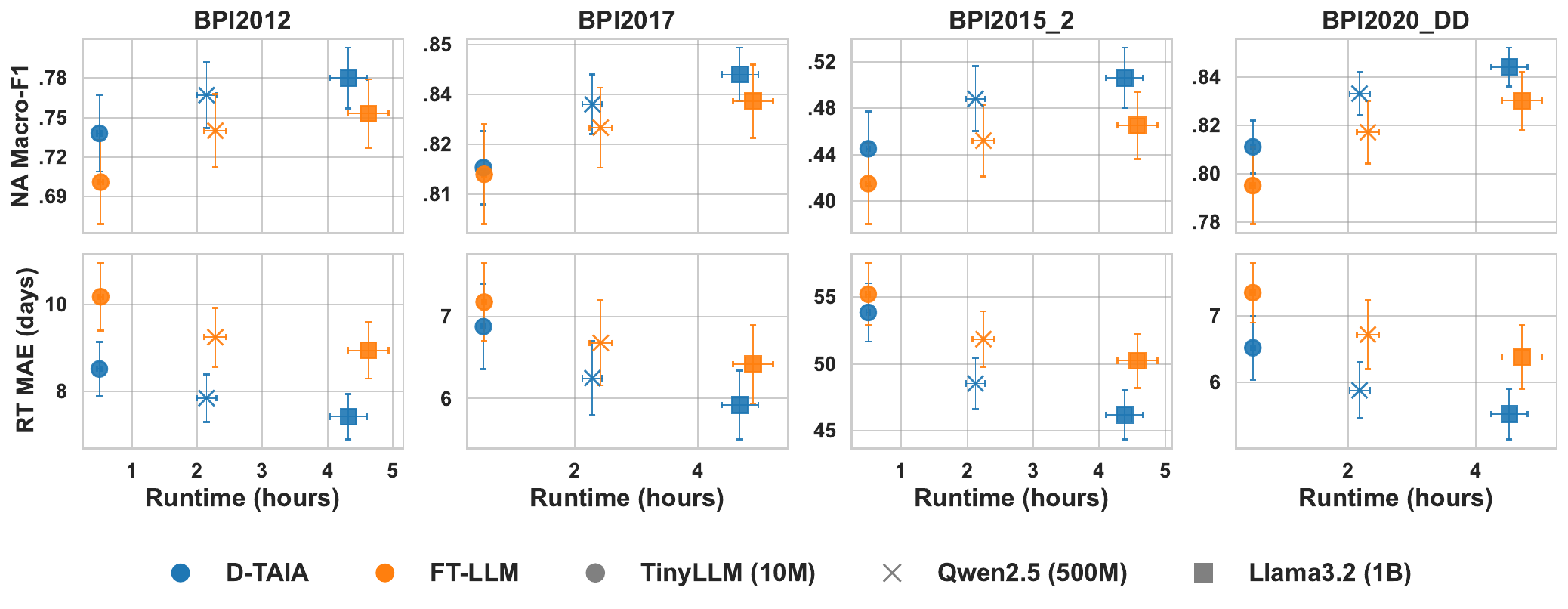}
    \vspace{-1em}
    \caption{D-TAIA achieves higher Macro-F1 score and lower MAE than FT-LLM~\cite{oyamada_2025} across all backbones and datasets.}
    \label{fig:backbone_vs_runtime}
    \vspace{-1em}
\end{figure}

We compare our framework against Oyamada et al.~\cite{oyamada_2025} using three backbones: Tiny-LLM~\footnote{\url{https://huggingface.co/arnir0/Tiny-LLM}} (10M parameters), Qwen2.5~\footnote{\url{https://huggingface.co/Qwen/Qwen2.5-0.5B}} (500M parameters) and Llama3.2~\footnote{\url{https://huggingface.co/meta-llama/Llama-3.2-1B}} (1B parameters). From the graphs in Fig.~\ref{fig:backbone_vs_runtime} two trends emerge. Scaling the backbone improves both metrics at the cost of runtime. D-TAIA's MAE is consistently lower than FT-LLM's at comparable runtime, with non-overlapping confidence intervals on BPI2012 and BPI2020\_DD across all three backbones. The two models are closer on BPI2017 and BPI2015\_2, where MAE intervals overlap throughout. D-TAIA's Macro-F1
score exceeds FT-LLM's at every backbone and dataset, though confidence intervals overlap at each individual comparison, so this should be read as a consistent small edge rather than a backbone-by-backbone win. The clearest scaling effect is on BPI2015\_2. Macro-F1 rises by 13.7\% from Tiny-LLM to Llama3.2, roughly two to three times the relative gain seen on the other three logs ($3$--$6\%$). This suggests that backbone capacity matters most on the highest-entropy log, while the gains from D-TAIA's architecture over FT-LLM stem from the added components rather than from added compute.

\vspace{-0.5em}
\subsubsection{Training Data Sensitivity.} 
\begin{figure}[t]
    \centering
    \includegraphics[width=1\linewidth]{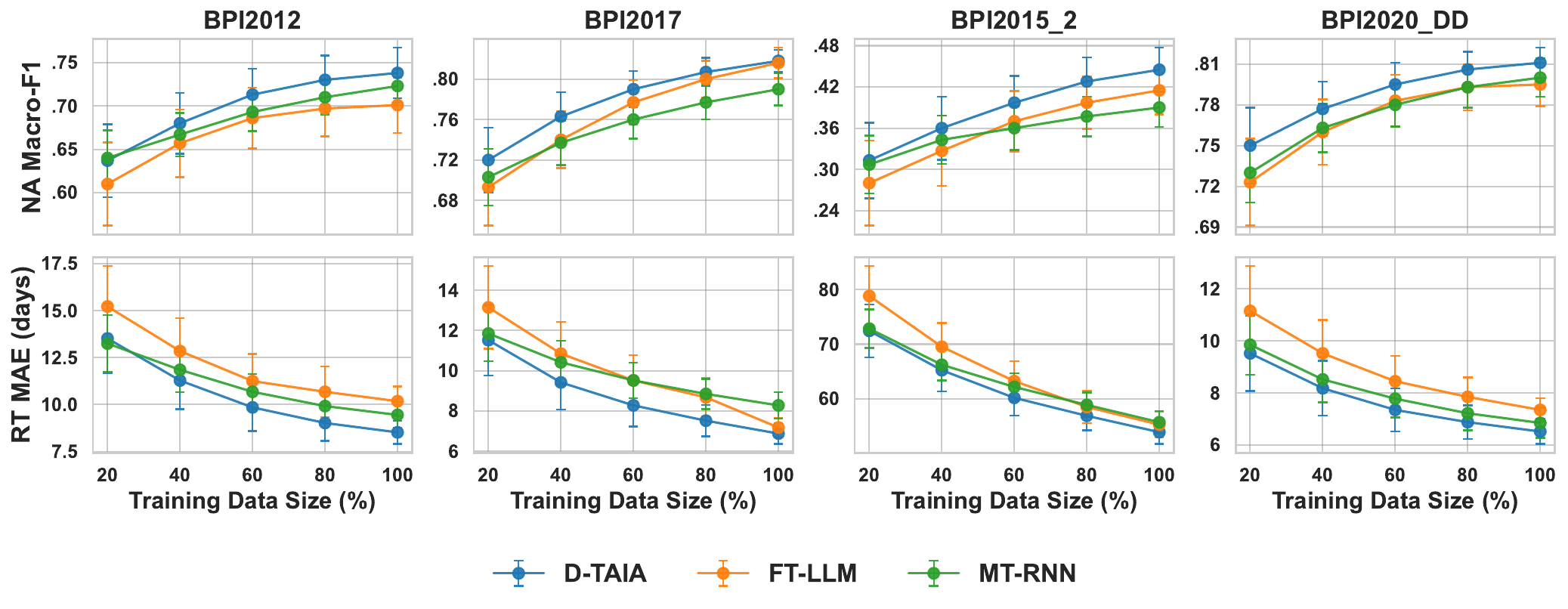}
    \vspace{-1em}
    \caption{D-TAIA attains the best point estimate for Macro-F1 and RT MAE at every training-data budget across all four logs. The advantage over FT-LLM is clearest and most persistent at full data on the high-entropy BPI2015\_2, while the three models converge most closely on BPI2017 and the near-deterministic BPI2020\_DD.}
    \label{fig:training_data}
        \vspace{-1em}
\end{figure}

D-TAIA attains the best point estimate for both Macro-F1 and RT MAE at every training-data budget across all four logs (cf. Fig.~\ref{fig:training_data}). At 20\% training data, confidence intervals for D-TAIA, FT-LLM, and MT-RNN overlap on each log, reflecting the greater estimation uncertainty at small sample sizes rather than an absence of effect. D-TAIA's advantage over FT-LLM is clearest and most persistent at full data on the high-entropy BPI2015\_2, while the three models converge most closely on BPI2017 and the near-deterministic BPI2020\_DD.

\subsubsection{Prefix Length Sensitivity.} 
\begin{figure}[h]
    \centering
    \includegraphics[width=1\linewidth]{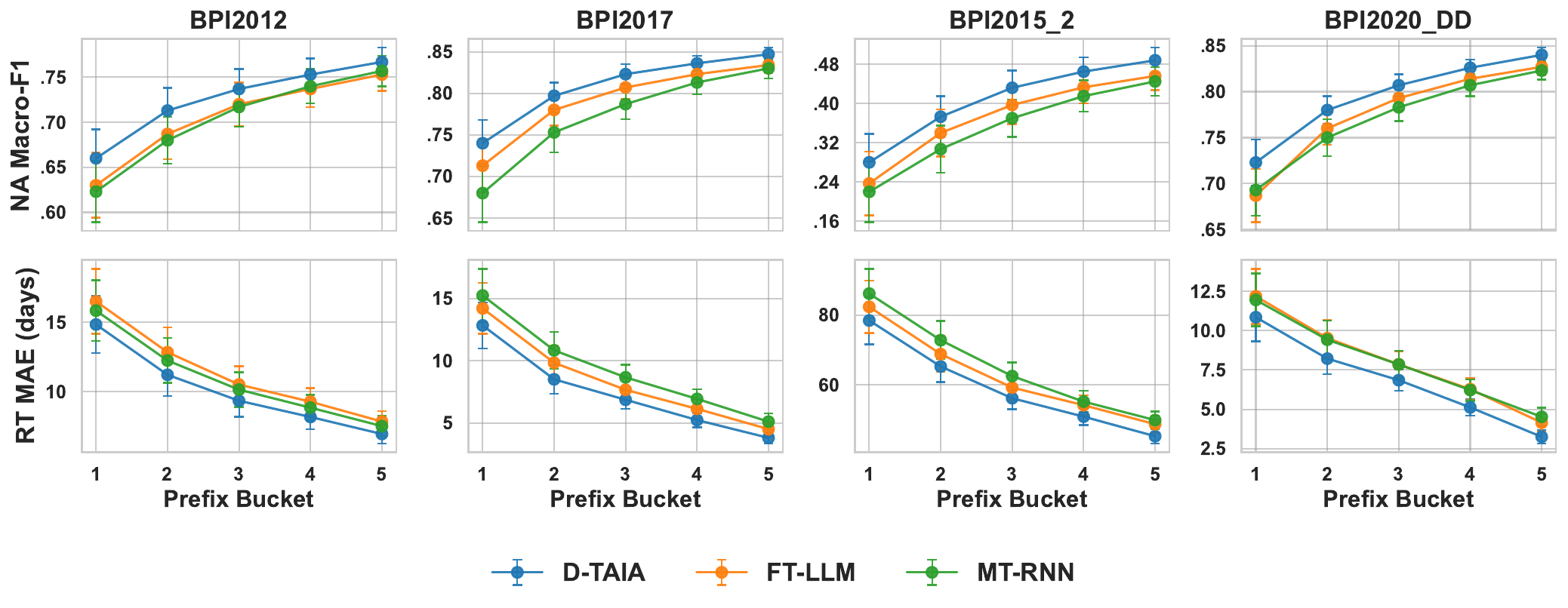}
    \vspace{-1em}
    \caption{D-TAIA's advantage over FT-LLM and MT-RNN in macro-F1 and RT MAE is largest at short prefixes (bucket 1) and narrows toward longer prefixes (bucket 5) across all four logs. The residual gap at bucket 5 remains proportionally largest on the high-entropy BPI2015\_2.}
    \label{fig:prefix_length}
    \vspace{-1em}
\end{figure}

Prefixes are grouped into five equal-width buckets by completion fraction, from bucket 1 (shortest) to
bucket 5 (longest) (cf. Fig.~\ref{fig:prefix_length}). D-TAIA's advantage over FT-LLM is largest at bucket 1, by 3.0--6.0 percentage points in Macro-F1, and narrows through bucket 5. The residual gap at bucket 5 remains proportionally largest on the high-entropy BPI2015\_2. RT MAE follows the same pattern. D-TAIA is lower than FT-LLM at every bucket, with the largest gap at bucket 1 and the smallest at bucket 5.

\vspace{-0.5em}
\subsubsection{Ablation Study.}
To isolate the contribution of each architectural component, four variants of D-TAIA are evaluated, each removing exactly one design decision while holding all others constant. BPI2020\_DD and BPI2015\_2 are used for their contrast in entropy. Table~\ref{tab:ablation_comparison} reports Macro-F1 and MAE for each
variant against the full model, with 95\% CIs and significance markers. Removing DATL produces the largest point-estimate drop on every metric and is the only variant whose 95\% CI is distinguishable from the full model on both Macro-F1 and MAE in the high-entropy setting, and on MAE in the low-entropy setting. However, this effect is not purely attributable to the loss of pre-trained structure. FAISS retrieval is not disabled when DATL is removed, but queries an encoder left at random initialization. This means that part of the degradation reflects retrieval against noise rather than the absence of retrieval itself. Removing FAISS shows the reverse pattern, hurting MAE more than Macro-F1 and is itself significant on high-entropy MAE. This indicates retrieval's contribution is concentrated in the harder, more data-scarce setting. Removing TAIA produces the smallest point-estimate effect of the four. This ablation contrasts entropy, not the distributional shift TAIA is designed to address. We leave a direct evaluation of TAIA under distributional shift to future work.
 
\begin{table}[t]
\centering
\scriptsize
\setlength{\tabcolsep}{7pt}
\renewcommand{\arraystretch}{0.95}
\caption{Ablation results of the framework's components. Bold marks the
best value per column. $^{*}$ marks a variant whose 95\% CI does not
overlap the full model's (D-TAIA) on that metric.}
\label{tab:ablation_comparison}
\vspace{-1.0em}
\resizebox{\textwidth}{!}{%
\begin{tabular}{lcc|cc}
\toprule
 & \multicolumn{2}{c|}{\textbf{High Entropy (BPI2015\_2)}} & \multicolumn{2}{c}{\textbf{Low Entropy (BPI2020\_DD)}} \\
\textbf{Variant} & Macro-F1 & MAE & Macro-F1 & MAE \\
\midrule
D-TAIA (full)           & $\mathbf{.445\pm.032}$ & $\mathbf{53.85\pm2.18}$ & $\mathbf{.811\pm.011}$ & $\mathbf{6.52\pm0.48}$ \\
-- DATL       & $.352\pm.048^{*}$ & $68.45\pm3.95^{*}$ & $.783\pm.018$ & $7.85\pm0.68^{*}$ \\
-- Domain ID & $.385\pm.042$ & $58.22\pm2.65$ & $.790\pm.016$ & $6.92\pm0.55$ \\
-- FAISS      & $.418\pm.036$ & $61.85\pm2.88^{*}$ & $.807\pm.012$ & $6.68\pm0.52$ \\
-- TAIA       & $.425\pm.034$ & $55.28\pm2.35$ & $.806\pm.013$ & $6.72\pm0.50$ \\
\bottomrule
\end{tabular}%
}
\vspace{-0.75em}
\end{table} 
\section{Conclusion and Future Work} 
\label{sec:conclusion-future_work}

This paper presented D-TAIA, a framework for joint NA and RT prediction under data scarcity and high process entropy. Three findings follow. First, FAISS-based retrieval addresses the regression weakness of autoregressive LLMs more effectively than with direct prediction. Second, combining DATL pre-training with attention-based inference is viable for joint PPM using only a 10M-parameter backbone. Third, D-TAIA matches or improves upon a fine-tuned LLM and an RNN baseline across four event logs, with the clearest margins at short prefix lengths and on the highest-entropy log. TAIA's inference strategy targets robustness to distributional shift by design, but we do not evaluate this directly and treat this as open. 

These findings suggest three directions for future work. TAIA's small ablation effect motivates testing it on a backbone where structural adaptation may matter more, such as a GNN encoding process topology directly. The FusionGate weight $\beta$ is fixed rather than learned. A gate conditioned on prefix entropy and retrieval confidence could shift the balance as information accumulates over a case. Finally, retrieval under a removed DATL stage still executes, only against an untrained embedding space. Injecting retrieved neighbor prefixes directly into the context window would test whether retrieval quality depends more on the backbone's pre-trained representations than on DATL itself.  

\bibliographystyle{splncs04}
\bibliography{sn-bibliography}

\end{document}